\documentclass{article}

\usepackage{microtype}
\usepackage{graphicx}
\usepackage{subfigure}
\usepackage{booktabs} % for professional tables

\usepackage{hyperref}
\usepackage{float}
\usepackage{adjustbox}
\usepackage{dblfloatfix}

\usepackage[accepted]{icml2023}

\usepackage{multirow}
\usepackage{amsmath}
\usepackage{amssymb}
\usepackage{mathtools}
\usepackage{amsthm}

\usepackage[capitalize,noabbrev]{cleveref}

\theoremstyle{plain}
\newtheorem{theorem}{Theorem}[section]

\newtheorem{lemma}[theorem]{Lemma}

\theoremstyle{definition}

\theoremstyle{remark}

\usepackage[textsize=tiny]{todonotes}

\icmltitlerunning{Distribution Free Prediction Sets for Node Classification}

\begin{document}

\twocolumn[
\icmltitle{Distribution Free Prediction Sets for Node Classification}




\begin{icmlauthorlist}
\icmlauthor{Jase Clarkson}{oxf}
\end{icmlauthorlist}

\icmlaffiliation{oxf}{Department of Statistics, University of Oxford}

\icmlcorrespondingauthor{Jase Clarkson}{jason.clarkson@stats.ox.ac.uk}

\icmlkeywords{Conformal Prediction, Node Classification, Distribution Free, Graph Neural Networks}

\vskip 0.3in
]



\printAffiliationsAndNotice{} % otherwise use the standard text.

\begin{abstract}
Graph Neural Networks (GNNs) are able to achieve high classification accuracy on many important real world datasets, but provide no rigorous notion of predictive uncertainty. Quantifying the confidence of GNN models is difficult due to the dependence between datapoints induced by the graph structure.

We leverage recent advances in conformal prediction to construct prediction sets for node classification in inductive learning scenarios. We do this by taking an existing approach for conformal classification that relies on \textit{exchangeable} data and modifying it by appropriately weighting the conformal scores to reflect the network structure. We show through experiments on standard benchmark datasets using popular GNN models that our approach provides tighter and better calibrated prediction sets than a naive application of conformal prediction. The code is available at \href{https://anonymous.4open.science/r/graph_cp-56EE/README.md}{this link}.
\end{abstract}

\section{Introduction}
Machine learning on graph structured data has seen a boom of popularity in recent years, with applications ranging from recommendation systems to biology and physics. Graph neural networks are quickly maturing as a technology; many state of the art models are commoditised in frameworks such as Pytorch Geometric \cite{pyg} and DGL \cite{dgl}. Despite their overwhelming popularity and success, very little progress has been made towards quantifying the uncertainty of the predictions made by these models, a vital step towards robust real world deployments.

In related areas of machine learning such as computer vision, conformal prediction \cite{vgs} has emerged as a promising candidate for uncertainty quantification \cite{angelopoulos2020sets}. Conformal prediction is a very appealing approach as it is compatible with any black box machine learning algorithm and dataset as long as the data is statistically exchangeable. The most wide-spread method, so called \textit{split-conformal}, also requires trivial computational overhead when compared to model fitting. 

% Conformal prediction uses the assumption that the test data have exchangeable statistical properties in order to assess the uncertainty of predictions.

Networks, or graph structured data is in general not exchangeable and so the guarantees provided by conformal prediction in its naive form do not hold. Recent work by \citet{cpbe} extends conformal prediction to the non-exchangeable setting and provides theoretical guarantees on the performance of conformal prediction in this setting. We leverage insights from \cite{cpbe} to apply conformal prediction in the node classification setting. The key insight is that for a homophilous graph, the model calibration should be similar in a neighbourhood around any given node. We leverage this insight to localise the calibration of conformal prediction. We show that our method improves calibration of predictive uncertainty and provides tighter prediction sets when compared with a naive application of conformal prediction across several state of the art models applied to popular node classification datasets.

This paper is structured as follows; we begin by reviewing related work in Section \ref{sec:rw}. In Section \ref{sec:cp} we give an introduction to the relevant background material on conformal prediction. We introduce our method for adapting conformal prediction to networks in Section \ref{sec:naps}. We then detail our experimental setup and provide marginal coverage statistics in Section \ref{sec:exp}, and discuss the conditional coverage properties of our method in {\color{black}{Section}}   \ref{sec:cc}. We provide an ablation study to better understand our results in \ref{sec:ablation} and finally give our conclusions and discuss avenues for future work in {\color{black}{Section}} \ref{sec:conc}.

\section{Related Work} \label{sec:rw}
There is no standard approach to estimating the predictive uncertainty of neural network models. Many works take a Bayesian approach \cite{bnn}, where the goal is to learn a distribution over the network weights and represent uncertainty via the posterior distribution. Bayesian approaches however become quickly computationally intractable for large models and datasets, which has lead to approximations of Bayesian learning such as Deep Ensembles \cite{dens} and Variational Inference \cite{vbayes}. These approximations too come with practical drawbacks, such as the need to explore distinct regions of the parameter space in Deep Ensembles. 

In the graph setting, several variants of Bayesian GNNs have been proposed \cite{gbnn, gbnn_mcmc, gbnn_ss}, all of which require modifications to either model architecture or the model training procedure. The methods introduced in this work are deployed \textit{after} model training and have trivial computational overhead compared to model fitting.

Conformal prediction \cite{vgs} has seen a surge in popularity in recent years, especially amongst the deep learning community \cite{stutzlearning, einbindertraining, teng2022predictive}. In the exposition below we will focus on conformal classification as introduced in \cite{aps} as that is the object of study in this work, but note that other approaches to conformal classification exist such as that introduced in \cite{cp_lamb}. Conformal prediction may also be used to construct prediction intervals for regression \cite{lei2018distribution, cqr} or to control more general risk functions \cite{c_risk}.  

While we are not aware of prior work extending conformal prediction to networks, very recent work has considered the application of conformal prediction to time series,
in which observations are in general not exchangeable. 
%which is also non-exchangeable in general. 
These algorithms often assume the distribution can shift over time, even adversarially, and as such utilise online learning \cite{gibbs2021adaptive, zaffran2022adaptive, faci} or game theoretic \cite{bastanipractical} techniques. In contrast, we assume that the data is non-exchangeable but that the adjacency matrix of the graph is indicative of the dependency structure between data points.

\section{Conformal Prediction} \label{sec:cp}
Conformal prediction is a family of algorithms that generate finite sample valid prediction intervals or sets from an arbitrary black box machine learning model. In this work we employ a convenient approach known as \textit{split} conformal prediction \cite{pap, lei2018distribution}. {\color{black} Split conformal prediction may be thought of as a "wrapper" around a fitted model that uses a set of exchangeable held out data to calibrate prediction sets.}  Amazingly, the predictive model does not even need be well specified for these guarantees to hold (although the prediction intervals or sets may not be useful in this case). An excellent tutorial is provided by \citet{cp_gentle}. 

\subsection{The Exchangeable Case}
In a $K-$class classification model suppose that we have a fitted model $\hat{f}: \mathcal{X} \rightarrow [0,1]^K$ that outputs the probability of each class. Given an exchangeable set of held-out calibration datapoints $\left(X_{1}, Y_{1}\right), \ldots,\left(X_{n}, Y_{n}\right)$ (held out meaning they were not used to fit the model) and a new evaluation point $\left ( X_{n+1}, Y_{n+1} \right )$, conformal prediction constructs a \textit{prediction set} {\color{black}$\widehat{C}_n(X_{n+1})$} that satisfies
\begin{equation}\label{eq:lower_upper}
    1-\alpha \leq \mathbb{P}\left(Y_{n+1} \in {\color{black}\widehat{C}_n\left(X_{ n+1 }\right)}\right) \leq 1-\alpha+\frac{1}{n+1}
\end{equation}
for a user specified error rate $\alpha \in [0, 1]$. Conformal prediction relies on a \textit{score function} $S: \mathcal{X} \times \mathcal{Y} \rightarrow \mathbb{R}$, which measures the calibration of the prediction at a given datapoint. Given a score function $S$, the procedure for constructing a prediction set is very simple; for each datapoint $\left(X_{i}, Y_{i}\right)$ in the calibration set, compute the score $s_{i}= S(X_i, Y_i)$. Define $1-\hat{\alpha}$ to be the $\lceil(n+1)(1-\alpha)\rceil / n$ empirical quantile of the scores $s_1, \dots, s_n$, and finally create the prediction set $$\widehat{C}_n\left(X_{n+1}\right)=\left\{y: S\left(X_{n+1}, y \right) \leq 1 - \hat{\alpha}\right\}.$$ 

A popular conformal prediction procedure for classification problems is known as Adaptive Prediction Sets (APS, \cite{aps}). To motivate the APS score function, suppose we have access to an \textit{oracle} classifier $\pi$ that exactly matches the true conditional distribution (i.e. $\pi(x) = \mathbb{P}(Y_{n+1} | X_{n+1} = x)$). Then to construct a $1-\alpha$ prediction set from the oracle, we simply sort the probabilities into descending order, and add labels to the set until the cumulative probability exceeds $1-\alpha$ (with appropriate tie breaking to ensure exact coverage). 

Let $$\pi_y(x)=\mathbb{P}(Y=y \mid X=x)$$ for all $y \in \mathcal{Y}$ and denote the (reverse) order statistics of the oracle classifier probabilities by $$\pi_{(1)}(x) \geq \pi_{(2)}(x) \geq \ldots \geq \pi_{(K)}(x).$$ For any $\tau \in [0,1]$ define the generalised conditional quantile as
\begin{multline}
    L(x ; \pi, \tau)=\min \{k \in\{1, \ldots, K\}: \\ \pi_{(1)}(x)+\pi_{(2)}(x)+\ldots+\pi_{(k)}(x) \geq \tau \}. \nonumber
\end{multline}

One can now define the set valued function
\begin{eqnarray*}
\mathcal{S}(x, u ; \pi, \tau)= 
\begin{cases}\text {Labels of the }\\ L(x ; \pi, \tau)-1 \text { largest } \pi_y(x), \\
\quad \quad \text { if } u \leq V(x ; \pi, \tau), \\ 
\\ \text {Labels of the }\\ L(x ; \pi, \tau) \text { largest } \pi_y(x), \\
\quad \quad \text { otherwise }\end{cases}
\end{eqnarray*}

where 
\begin{equation*}
V(x ; \pi, \tau)=\frac{1}{\pi_{(L(x ; \pi, \tau))}(x)}\left[\sum_{c=1}^{L(x ; \pi, \tau)} \pi_{(c)}(x)-\tau\right] .
\end{equation*}

The oracle prediction set is then be defined as
$$
C_\alpha^{\text {oracle }}(x)=\mathcal{S}(x, U ; \pi, 1-\alpha)
$$
where $U \sim$ Uniform(0,1) is independent of everything else. The above is saying one should break ties proportional to the gap between the cumulative sum of the ordered probabilities until the true label is included and the desired level $\tau$. 

% Let $\lbrace \pi_{(1)}, \dots, \pi_{(K)} \rbrace$ be the order statistics of the conditional probabilities $\pi(x)$ {\color{black}{so that $\pi_{(1)} \ge \pi_{(2)} \ge \cdots \ge \pi_{(K)}$}}. Prediction sets can be constructed from the oracle as
%     $$\left\{\pi_{(1)}, \ldots, \pi_{(k)}\right\}, \text{where} \;  k=\inf \left\{k^{\prime}: \sum_{j=1}^{k^{\prime}} \pi_{(j)} \geq 1-\alpha\right\}.$$
{\color{black}In practice the probabilities given by a fitted classifier $\hat{f}(x)$ will usually not be exactly equal to $\mathbb{P}(Y_{n+1} | X_{n+1}=x)$. APS instead measures the deviation from the oracle procedure required to achieve the desired level of coverage on the calibration data; the conformal score is defined as}
\begin{equation} \label{eq:aps_score}
S(X_i, Y_i) = \min \lbrace \tau \in [0,1] : Y_i \in \mathcal{S}(X_i, U_i; \hat{f}, \tau) \rbrace.
\end{equation}
where again, $U_i \sim U[0,1]$, independent of everything else.
% Intuitively, labels are added to the set until the true label is included, and the score is the cumulative probability of this set. To give an example, if we ran this procedure on a set of calibration data and found that we needed to use the level $94\%$ to get $90\%$ coverage, then we would use $\hat{q} = 0.94$ as the threshold when constructing new prediction sets. 
{\color{black}To give a concrete example, suppose we want to construct prediction sets that contain the true label $90\%$ of the time (so $\alpha = 0.1$). It could be the case that, on our held out data, if we simply add up the ordered softmax outputs until their cumulative sum exceeds $0.9$ we actually get $85\%$ coverage, due to the model being miss-calibrated. Using APS we might calculate that if we construct prediction sets using $1 - \hat{\alpha} = 0.94$ we get $90\%$ coverage, and by exchangeability this translates to $90\%$ coverage on any new test point. We would therefore use the level $\hat{\alpha} = 0.06$ to construct our new prediction sets. 
\vspace{-2mm}
\subsection{Beyond Exchangeability}
Conformal prediction in the form presented above relies on the assumption that the data points $Z_{i}=\left(X_{i}, Y_{i}\right)$ are exchangeable. 
% \begin{enumerate}
%     \item The algorithm $\mathcal{A}$, mapping data sets to fitted models $\hat{f}$, is assumed to treat the data points symmetrically, in the sense that 
%     \begin{equation}
%     \mathcal{A}\left(\left(x_{\pi(1)}, y_{\pi(1)}\right), \ldots,\left(x_{\pi(n)}, y_{\pi(n)}\right)\right)=\mathcal{A}\left(\left(x_{1}, y_{1}\right), \ldots,\left(x_{n}, y_{n}\right)\right)
%     \end{equation}
%     for all permutations $\pi$ on $\lbrace 1, \dots, n \rbrace$. 
%     \item The data $Z_{i}=\left(X_{i}, Y_{i}\right)$ are assumed to be exchangeable. 
% \end{enumerate}
The exchangeable form of conformal prediction provides no guarantee if these assumptions are violated, however \textit{non-exchangeable conformal prediction} was introduced in the pioneering work of \citet{cpbe}.

Formally, the non-exchangeable conformal prediction procedure assumes a choice of deterministic fixed weights $w_1, \dots, w_n \in [0,1]$ (normalized as detailed in \cite{cpbe}). 
%and normalises them as
% \begin{equation} \label{eq:wnorm}
% \tilde{w}_{i}=\frac{w_{i}}{w_{1}+\cdots+w_{n}+1}, i=1, \ldots, n \text {, and } \tilde{w}_{n+1}=\frac{1}{w_{1}+\cdots+w_{n}+1}.
% \end{equation}
As before, one computes the scores $s_1, \dots, s_n$ but now defines the prediction set in terms of the \textit{weighted quantiles} of the score distribution 
% TODO: Fix this.
\begin{eqnarray}
\lefteqn{\widehat{C}_{n}\left(X_{n+1}\right)} \nonumber \\
&=&\left\{ y \in \mathcal{Y}: S\left(X_{n+1}, y\right) \leq \right. \nonumber \\  && \left. \mathrm{Q}_{1-\alpha}\left(\sum_{i=1}^{n} w_i \cdot \delta_{s_i} + w_{n+1} \cdot \delta_{+\infty}\right) \right\} %\rbrace
    \label{eq:wquant}
\end{eqnarray} 
where $\mathrm{Q}_{\tau}(\cdot)$ denotes the $\tau$-quantile of a distribution {\color{black} and $\delta_x$ denotes a point mass at $x$.} 

Non-exchangeable conformal prediction also comes with performance guarantees; the authors define the \textit{coverage gap} 
\begin{equation} \label{eq:cg}
\text { Coverage gap }=(1-\alpha)-\mathbb{P}\left\{Y_{n+1} \in \widehat{C}_{n}\left(X_{n+1}\right)\right\}
\end{equation}
as the loss of coverage when compared to the exchangeable setting, and show that this can be bounded as follows: let $Z=\left(\left(X_{1}, Y_{1}\right), \ldots,\left(X_{n+1}, Y_{n+1}\right)\right)$ be the full dataset and define $Z^i$ as the same dataset after swapping the test point and the $i^{th}$ training point
\begin{multline}
    Z^{i}=\left(\left(X_{1}, Y_{1}\right), \ldots,\left(X_{i-1}, Y_{i-1}\right),\left(X_{n+1}, Y_{n+1}\right), \right.  \\ \left. \left(X_{i+1}, Y_{i+1}\right), \ldots,\left(X_{n}, Y_{n}\right),\left(X_{i}, Y_{i}\right)\right)).  \nonumber
\end{multline}

Then the coverage gap in Equation \eqref{eq:cg} can be bounded as (Theorem 2a, \citet{cpbe}):
\begin{equation} \label{eq:gapbound}
\text { Coverage gap } \leq \frac{\sum_{i=1}^{n} w_{i} \cdot \mathrm{d}_{TV}\left(Z, Z^{i}\right)}{1+\sum_{i=1}^{n} w_{i}}
\end{equation}
where $\mathrm{d}_{TV}$ is the total variation distance. To make this bound small one would like to place a large weight $w_i$ on datapoints that are drawn from a similar distribution to the test point $(X_{n+1}, Y_{n+1})$. 

% \begin{table*}[t!]
% 	\centering
% 	\caption{Statistics for the Flickr, Reddit2 and Amazon Computers datasets, with notation as in Subsection \ref{subsec:data}.}
% 	\label{tab:datasets}
% 	\begin{tabular}{cccccccccc}
% 		\toprule
% 		  Dataset & Nodes & Edges & \# Feat & \# Classes & \# Test Nodes & $\left | \mathcal{N}^{cal} \right | $ & $\hat{H}$ & $H_{rand}$\\
% 		\midrule
% 		Flickr & 89,250  & 899,756 & 500 & 7 & 22313 & 5161 & 0.319 & 0.266 \\
% 		Reddit2 & 232,965  & 23,213,838 & 602 & 41 & 55334 & 22160 & 0.812 & 0.051 \\
%         Amazon Comp. & 13752 & 491,722 & 767 & 10 & 12000 & 11033 & 0.785 & 0.208 \\
% 		% C & 0.2716  & 0.1826 & 0.2836 & 0.1836 \\
% 		\bottomrule
% 	\end{tabular}
    
% \end{table*}

\section{Conformal Prediction for Node Classification} \label{sec:naps}
Consider now the node classification setting: we are given a graph $G=(V,E)$, and for each node $i \in V$ we are given a node feature vector $X_i \in \mathbb{R}^F$ and a label $Y_i \in \mathcal{Y}$. A standard pipeline for node classification usually consists of a GNN model that produces a node embedding $h_i \in \mathbb{R}^H$ followed by a classifier $f: \mathbb{R}^H \rightarrow \mathcal{Y}$. 

Here the data points $Z_i = (X_i, Y_i)$ are certainly not assumed to be exchangeable; the underlying principle of GNN models is that the adjacency matrix {\color{black}{of $G$}} provides information about the dependency between datapoints (and hence neighbourhood information  {\color{black}{of $G$}}  is aggregated and used for prediction). \citet{cpbe} show in particular that non-exchangeable data can be navigated when the fitted model is a symmetric function of the test data. Our method is based on the observation that using only training data to fit the model trivially satisfies this assumption. In particular, this excludes the transductive setting.

% Assuming we use a permutation invariant GNN architecture for the embedding step, this pipeline satisfies the first condition (to see this, augment each $Z_i$ with the $i^{th}$ row of the adjacency matrix). However the datapoints are in general not exchangeable and in fact heavily influenced by the adjacency matrix of the graph. 

We combine non-exchangeable conformal prediction with the information given by the adjacency matrix to produce an algorithm for constructing prediction sets for node classification, which we call Neighbourhood Adaptive Prediction Sets (NAPS). The first variant simply localises the calibration to a neighbourhood of the network;  we set the weights in Equation \eqref{eq:wquant} to $w_i = 1$ if $i \in \mathcal{N}^K_{n+1}$, where $\mathcal{N}^K_{n+1}$ is the $K$-hop neighbourhood of node $n+1$. We then apply non-exchangeable conformal prediction with the APS scoring function in Equation \eqref{eq:aps_score}.

\begin{figure}[b!]
\vspace{0.5cm}
    \centering
    \includegraphics[scale=0.8
    ]{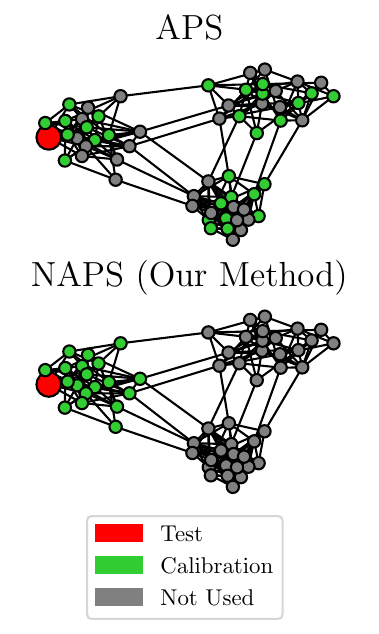}
    \caption{An illustration of the nodes used for calibrating conformal prediction via an "out the box" application of APS (top panel), which randomly splits the data, and the nodes used in NAPS (bottom panel). NAPS localises the calibration nodes to a neighbourhood of the test node.}
    \label{fig:my_label}
\end{figure}

The coverage gap of NAPS is bounded as
\begin{equation} \label{eq:graphgapbound}
\text { Coverage gap } \leq \frac{\sum_{i\in \mathcal{N}^{k}_{n+1}} \mathrm{d}_{TV}\left(Z, Z^{i}\right)}{1+ | \mathcal{N}^{k}_{n+1} |}
\end{equation}
by simple substitution into Equation \eqref{eq:gapbound}. This bound will be small if the $k$-hop neighbours of node $n+1$ are distributed similarly, which is otherwise known as \textit{homophily} \cite{homo}. 

Homophily is a key principle of many real world networks, where linked nodes often belong to the same class and have similar features, and is in crucial for good performance in many popular GNN architectures (although recent work has considered the heterophilic case, see \cite{heterognn}, which we will discuss in the future work section). This is also related to network \textit{homogeneity}, where nodes in a neighbourhood play similar roles in the network and are considered interchangeable on average. 

Motivated by the principle of homophily, we also explore two variants of NAPS that place more weight on closer neighbours than those further away. Formally we introduce a weighting function $w(k)$ setting $w_i = w(k)$ for a node $i$ that is $k$ hops from the test node $n+1$. In this setting, letting $\mathcal{N}^{k'}_{n+1}$ represent the nodes that are exactly $k$ hops from node $n+1$, {\color{black}{similarly to \eqref{eq:graphgapbound}}} we have the coverage gap bound
\begin{equation} \label{eq:linearcg}
    \text { Coverage gap } \leq \frac{\sum_{k=1}^{K} w(k) \sum_{i\in \mathcal{N}^{k'}_{n+1}} \mathrm{d}_{TV}\left(Z, Z^{i}\right)}{1+ \sum_{k=1}^K w(k) | \mathcal{N}^{k'}_{n+1} | }.
\end{equation}

The first variant, which we call NAPS-H, uses a hyperbolic decay rate $w(k) = k^{-1}$, and the second, which we call NAPS-G, uses a geometric decay rate $w(k) = {2^{-k}}$.

{\color{black} Note that NAPS is not applicable in the transductive setting, as the fitted model $\hat{f}$ would depend on the node features in the test set, hence the conformal scores would no longer be exchangeable. It is however applicable in inductive settings where either the test set consists of multiple new graphs, or new nodes are added to an existing network. 

The neighbourhood depth parameter $K$ introduces a tradeoff; expanding the neighbourhood increases the sample size for calibration, but introduces nodes that may be progressively less topologically similar. In the form presented here we recommend only applying NAPS to large homophilous networks with dense 1 or 2 hop neighbourhoods, but we will discuss extensions in future work.}
% This approach could also modified to handle \textit{heterophilic} networks by setting $w_i = 1 - A_{i, n+1}$. 
{\color{black} 
\section{A Case Study: The Stochastic Block Model}

\begin{table*}[t!]
	\centering
	\caption{Statistics for the Flickr, Reddit2 and Amazon Computers datasets, with notation as in Subsection \ref{subsec:data}.}
	\label{tab:datasets}
	\begin{tabular}{cccccccccc}
		\toprule
		  Dataset & Nodes & Edges & \# Feat & \# Classes & \# Test Nodes & $\left | \mathcal{N}^{cal} \right | $ & $\hat{H}$ & $H_{rand}$\\
		\midrule
		Flickr & 89,250  & 899,756 & 500 & 7 & 22313 & 5161 & 0.319 & 0.266 \\
		Reddit2 & 232,965  & 23,213,838 & 602 & 41 & 55334 & 22160 & 0.812 & 0.051 \\
        Amazon Comp. & 13752 & 491,722 & 767 & 10 & 12000 & 11033 & 0.785 & 0.208 \\
		% C & 0.2716  & 0.1826 & 0.2836 & 0.1836 \\
		\bottomrule
	\end{tabular}
    
\end{table*}

To gain further insight into when we might expect NAPS to outpeform APS, we perform a theoretical study inspired by the Stochastic Block Model (SBM), a parametric model from network science. Suppose we have a graph on $2n$ nodes partitioned into two distinct groups, $C_1$ and $C_2$. We introduce two parameters, $p_{in}$, the probability of connecting two nodes in the same group, and $p_{out}$, the probability of connecting two nodes in different groups (usually $p_{out} << p_{in}$). We assume for nodes $i \in C_1$ that $(X_i, Y_i) \sim \pi_1$ i.i.d for all $i \in C_1$ where $\pi_1$ is a probability distribution, and that for nodes $j \in C_2$ that $(X_j, Y_j) \sim \pi_2$ i.i.d where $\pi_2 \neq \pi_1$, a different probability distribution. Importantly, we assume the group membership of each node is not observable, only implicit in the model specification.
 
Assume we have already fit a classifier $\hat{f}$ on a distinct training set, and our test network follows the block-model structure described above. The goal is to construct prediction sets for each node in the test set. If the group membership for each node was known, the problem would be easy; one could simply calibrate amongst only nodes in the same community. These nodes are drawn i.i.d from the same probability distribution, hence are exchangeable and so the standard guarantees for conformal prediction hold. As community membership is not observable, one must use the edges to determine whether two nodes are likely to be in the same group or not. Intuitively if $p_{in} >> p_{out}$ then on average linked nodes are much more likely to be in the same community, hence if there are enough neighbours to ensure sufficient sample size one would want to calibrate amongst the neighbours. 

We now provide a theoretical result quantifying this intuition, the proof of which is given in Appendix \ref{app:proof}.

\vspace{-1mm}
\begin{lemma} \label{lemma:1}
Assume the test data has the block model structure described above. Let $CG_{\text{APS}}$ be the coverage gap for prediction sets constructed using APS (i.e. calibrating using all available nodes), and $CG_{\text{NAPS}}$ be the coverage gap attained by the unweighted variant of NAPS calibrated amongst the 1-hop neighbours of each test node. Then $\mathbb{E}[CG_{\text{NAPS}}] < \mathbb{E}[CG_{\text{APS}}]$ if
\begin{equation} \label{eq:bdfull}
\mathbb{E} \left [ \frac{N_{out}}{N_{out} + N_{in} + 1}  \right ] < \frac{1}{2}
\end{equation}
\end{lemma}
}
where $N_{in} \sim Bin(n-1, p_{in})$, $N_{out} \sim Bin(n, p_{out})$ are Binomial random variables.

Note if we mean field approximate the expectation in Equation \eqref{eq:bdfull} we obtain a more intuitive expression in terms of the data generating parameters, namely that $\mathbb{E}[CG_{\text{NAPS}}] < \mathbb{E}[CG_{\text{APS}}]$ (approximately) when 
$$     \frac{np_{out}}{np_{out} + (n-1)p_{in} + 1} < \frac{1}{2}.
$$.

\section{Experiments} \label{sec:exp}
We now perform experiments with popular real world datasets and models to evaluate the performance of our procedure. Our experiments follow the following format: we split each graph into training, validation and test nodes (where the validation and test nodes are not available during model fitting i.e. an inductive node split). The training and validation sets are used for model fitting, and the test set is used to evaluate the conformal prediction procedure by constructing prediction sets and evaluating the empirical coverage. 
{\color{black}{Details of the implementation are found in Appendix \ref{app:training}.}}
{\color{black} 
\subsection{Evaluating Conformal Prediction} \label{app:eval_cp}
The observed coverage in a single application of conformal prediction is a \textit{random} quantity, where the randomness comes from the choice of which data points are used for calibration as well as the finite sample size of the calibration set (corresponding to the upper bound in Equation \eqref{eq:lower_upper}). It is therefore important to pick a large enough number of calibration points, and also repeat the experiment many times with different calibration/evaluation splits.

For simplicity we follow the guidelines given in \citet{cp_gentle}, which suggest using at least 1000 validation points, and we repeat each experiment 100 times with a different calibration/evaluation split; with this setup by the law of large numbers the probability of observing significant deviations from the true coverage is extremely low, %{\color{magenta}{be careful; that will depend on network size; they seem to consider an infinite 
and therefore we can evaluate the performance of our method with high confidence.

Conformal prediction in the exchangeable setting is usually deployed by splitting the data into a calibration set and an evaluation set. The calibration points are used to estimate the quantile of the score distribution, which is used to construct prediction sets for each evaluation point. In our setting, this corresponds to selecting the calibration and evaluation nodes randomly, which ignores the graph structure. The goal of our experimental setup is to study the improvement in the performance of conformal prediction when the graph structure is taken into account. 
}

\begin{table*}[t!]	

	\centering
	\caption{The test accuracy, empirical coverage, average prediction set size and average prediction set size conditional on coverage for all models considered on the Reddit2, Flickr and Amazon Computers datasets with $\alpha = 0.1$. Each column shows the median-of-means computed over 100 repetitions of the experiment, where each experiment is evaluated on a set of 1000 nodes. Bold indicates the best performing method.}
 \label{tab:size_coverage}
\begin{adjustbox}{max width=\textwidth}
	\begin{tabular}{ccccccccccccccc}
		\toprule
		\multirow{2.5}{*}{Dataset} & \multirow{2.5}{*}{Model} & Accuracy &  \multicolumn{4}{c}{Coverage} & \multicolumn{4}{c}{Size} & \multicolumn{4}{c}{Size $|$ Coverage}  \\
		\cmidrule{3-3} \cmidrule(lr){4-7} \cmidrule(lr){8-11} \cmidrule(lr){12-15} &
		& Top-1 & APS & NAPS & NAPS-H & NAPS-G & APS & NAPS & NAPS-H & NAPS-G & APS & NAPS & NAPS-H & NAPS-G \\
		\midrule
		\multirow{4}{*}{Reddit2} & GS-Mean & $0.914$ & $0.928$ & $0.895$ & $0.896$ & $0.899$ & $2.02$ & $\mathbf{1.59}$ & $\mathbf{1.59}$ & $\mathbf{1.59}$ & $2.12$ & $\mathbf{1.72}$ & $1.73$ & $1.73$ \\
		& GS-Max & $0.771$ & $0.918$ & $0.903$ &$0.904$ & $0.906$ & $3.97$& $\mathbf{3.20}$ & $\mathbf{3.20}$ & $3.21$ & $3.82$ & $\mathbf{3.30}$ & $3.31$ & $3.31$ \\
        & SD-SAGE & $0.844$ & $0.925$ & $0.896$ & $0.899$ & $0.897$ & $2.08$ & $1.64$ & $\mathbf{1.63}$ & $1.64$ & $2.11$ & $1.69$ & $\mathbf{1.67}$ & $1.68$ \\
        & SD-GCN & $0.827$ & $0.927$ & $0.896$ & $0.898$ & $0.899$ & $2.11$ & $1.66$ & $\mathbf{1.64}$ & $1.65$ & $2.14$ & $1.73$ & $\mathbf{1.70}$ & $1.71$ \\
		\midrule
		\multirow{4}{*}{Flickr} & GS-Mean & $0.503$ & $0.915$ & $0.904$ & $0.909$ & $0.910$ & $4.22$ & $\mathbf{4.01}$ & $4.10$ & $4.12$ & $4.26$ & $\mathbf{4.06}$ & $4.17$ & $4.19$ \\
		& GS-Max & $0.501$ & $0.910$ & $0.903$ & $0.907$ & $0.909$ & $4.32$ & $\mathbf{4.11}$ & $4.21$ & $4.23$ & $4.34$ & $\mathbf{4.14}$ & $4.24$ & $4.27$  \\
        & SD-SAGE & $0.500$ & $0.912$ & $0.904$ & $0.907$ & $0.908$ & $4.27$ & $\mathbf{4.07}$ & $4.15$ & $4.17$ & $4.31$ & $\mathbf{4.11}$ & $4.21$ & $4.23$ \\
		& SD-GCN & 0.496 & $0.912$ & $0.903$ & $0.908$ & $0.909$ & $4.28$ & $\mathbf{4.09}$ & $4.18$ & $4.20$ & $4.33$ & $\mathbf{4.12}$ & $4.22$ & $4.25$ \\
        \midrule
        \multirow{4}{*}{Amazon} & GS-Mean & $0.854$ & $0.905$ & $0.902$ & $0.902$ & $0.904$ & $1.50$ & $\mathbf{1.44}$ & $\mathbf{1.44}$ & $1.46$ & $1.57$ & $1.50$ & $\mathbf{1.49}$ & $1.52$ \\
		& GS-Max & $0.765$ & $0.902$ & $0.903$ & $0.902$ & $0.902$ & $2.15$ & $\mathbf{1.98}$ & $2.01$ & $2.01$ & $2.18$ & $\mathbf{2.04}$ & $2.06$ & $2.07$ \\
        & SD-SAGE & $0.815$ & $0.912$ & $0.905$ & $0.905$ & $0.906$ & $1.77$ & $\mathbf{1.66}$ & $1.67$ & $1.67$ & $1.80$ & $\mathbf{1.72}$ & $1.74$ & $1.73$ \\
        & SD-GCN & $0.822$ & $0.911$ & $0.904$ & $0.905$ & $0.904$ & $1.75$ & $1.64$ & $\mathbf{1.63}$ & $1.64$ & $1.82$ & $1.72$ & $\mathbf{1.71}$ & $1.73$  \\
		\bottomrule
	\end{tabular}
\end{adjustbox}
\end{table*}

\subsection{Experimental Setup} \label{sec:e_setup}
In each experiment, we sample a batch of evaluation nodes and construct a $1-\alpha$ probability prediction set {\color{black} for each evaluation node} using NAPS as described in Section \ref{sec:naps}, as well as using a naive application of APS calibrated among all the nodes {\color{black} not in the evaluation set}. We then report the empirical coverage, average prediction set size and average size of the prediction set given that the set contains the true label across all nodes. 

For each experiment we sample 1000 nodes randomly from the nodes in the test set, and we perform 100 repetitions of the experiment {\color{black}(see Appendix \ref{app:eval_cp} for a justification of this approach)}. We only apply our method to large connected components from the test set following the discussion in Section \ref{sec:naps} (see Appendix \ref{app:dataset} for details on the datasets and the test set construction procedure).

\begin{table*}[b]
\caption{The SSCV and PCCV (as described in Equations \eqref{eq:sscv} and \eqref{eq:NCCV}) for all models considered on the Reddit2, Flickr and Amazon Computers datasets with $\alpha = 0.1$. Bold indicates the best performing method. All the NAPS variants use depth parameter $K=2$. }
    \begin{tabular}{cccccccccc}
        \toprule
		\multirow{2.5}{*}{Dataset} & \multirow{2.5}{*}{Model} &  \multicolumn{4}{c}{SSCV} & \multicolumn{4}{c}{PCCV} \\
		 \cmidrule(lr){3-6} \cmidrule(lr){7-10}&
		& APS & NAPS & NAPS-H & NAPS-G &  APS & NAPS & NAPS-H & NAPS-G \\
  		\midrule
        \multirow{4}{*}{Reddit2} & GraphSAGE-Mean & $0.087$ & $0.074$ & $\mathbf{0.072}$ & $0.074$ & $0.084$ & $\mathbf{0.052}$ & $0.054$ & $0.054$ \\
        & GraphSAGE-Max & $0.089$ & $0.075$ & $\mathbf{0.073}$ & $0.076$ & $0.083$ & $0.052$ & $\mathbf{0.051}$ & $0.053$ \\ 
        & ShaDow-SAGE & $0.075$ & $0.056$ & $\mathbf{0.053}$ & $0.055$ & $0.087$ & $0.054$ & $\mathbf{0.051}$ & $0.055$ \\
        & ShaDow-GCN & $0.078$ & $0.056$ & $\mathbf{0.054}$ & $0.056$ & $0.089$ & $0.053$ & $\mathbf{0.052}$ & $0.053$\\
        \midrule
        \multirow{4}{*}{Flickr} & GraphSAGE-Mean & $0.113$ & $\mathbf{0.064}$ & $0.074$ & $0.079$ & $0.080$ & $\mathbf{0.059}$ & $0.065$ & $0.065$ \\
        & GraphSAGE-Max & $0.102$ & $\mathbf{0.062}$ & $0.073$ & $0.075$ & $0.084$ & $\mathbf{0.061}$ & $0.067$ & $0.067$ \\ 
        & ShaDow-SAGE & $0.100$ & $\mathbf{0.054}$ & $0.085$ & $0.087$ & $0.089$ & $\mathbf{0.064}$ & $0.066$ & $0.067$ \\
        & ShaDow-GCN & $0.105$ & $\mathbf{0.054}$ & $0.079$ & $0.081$ & $0.091$ & $\mathbf{0.062}$ & $0.069$ & $0.071$ \\
        \midrule
        \multirow{4}{*}{Amazon} & GraphSAGE-Mean & $0.082$ & $0.073$ & $\mathbf{0.072}$ & $0.079$ & $0.076$ & $0.046$ & $0.042$ & $\mathbf{0.041}$ \\
        & GraphSAGE-Max & $0.081$ & $\mathbf{0.072}$ & $\mathbf{0.072}$ & $0.074$ & $0.078$ & $0.045$ & $\mathbf{0.042}$ & $\mathbf{0.042}$\\ 
        & ShaDow-SAGE & $0.065$ & $0.057$ & $0.058$ & $\mathbf{0.052}$ & $1.03$ & $0.056$ & $0.053$ & $\mathbf{0.050}$ \\
        & ShaDow-GCN & $0.066$ & $0.055$ & $\mathbf{0.054}$ & $0.055$ & $0.99$ & $0.051$ & $0.052$ & $\mathbf{0.049}$ \\
        \bottomrule
    \end{tabular} \label{tab:nccv}
\end{table*}

\subsection{Datasets and Models}\label{subsec:data}

We apply our method to {\color{black}three} popular node classification datasets, namely Reddit2 and Flickr introduced in \cite{graphsaint-iclr20} {\color{black}and Amazon Computers introduced in \cite{pitfalls}}. We apply two variants of two popular GNN models, namely GraphSAGE \cite{sage} with the mean and max aggregators, and the ShaDow \cite{shaDow} subgraph sampling scheme with GraphSAGE and GCN \cite{gcn} layers. 

NAPS relies on node homophily to minimize the coverage gap bound in Equation \eqref{eq:gapbound}. Here we verify here that each of these networks is homophilous. We measure this via the node homophily ratio defined in \citep{geom_gcn} as 
$$ H = \frac{1}{|\mathcal{V}|} \sum_{v \in \mathcal{V}} \frac{ | \{ (w,v) : w \in \mathcal{N}(v) \wedge y_v = y_w \} |  } { |\mathcal{N}(v)| }.$$

We define a homophilous network as one that has node homophily ratio larger than expected under a random assignment of labels. For a network with $K$ classes, assume each node is assigned class $k$ independently with probability $p_k$. Then for any $(v,w) \in \mathcal{V}$, we have
$$ \mathbb{P}(y_v = y_w) = \sum_{k=1}^K p_k^2. $$

It follows that the expected homophily ratio under random class assignment is
$$ \mathbb{E}[H] = \frac{1}{|\mathcal{V}|} |\mathcal{V}| \sum_{k=1}^K p^2_k = \sum_{k=1}^K p^2_k.$$

In Table \ref{tab:datasets} we report both the observed homophily ratio $\hat{H}$ computed over the induced subgraph of the test nodes for each network as well as the expected node homophily under a random assignment of the labels $H_{rand}$, using the relative node label frequencies as the probabilities $p_k$. %{\color{magenta}{In that table explain $\mathcal{N}^{cal}$ as the number of nodes in the largest connected component}}} 
We see that Reddit2 and Amazon Computers are strongly homophilous, while Flickr is relatively weakly so. 

The results for each dataset are displayed in Table \ref{tab:size_coverage}. We see across all models on all three datasets, NAPS produces well calibrated, tight prediction sets, while the naive application of APS tends to overcover and produces wider prediction sets. The outperformance of NAPS over APS on the Flickr dataset shows that strong homophily is not required. We also see that this outperformance persists over a range of different GNN architectures. While the three NAPS variants perform similarly on Reddit2 and Amazon, NAPS outperforms the weighted variants by a significant margin on the Flickr dataset. This is likely due to the relatively weak node homophily ratio, implying that the increase in effective sample size is more valuable than down-weighting more distant neighbours. 

\section{Conditional Coverage and Network Topology} \label{sec:cc}
In this section we will examine the conditional coverage properties of NAPS via two metrics, one generic and one specific to the graph setting. A set-valued predictor $\mathcal{C}: \mathcal{X} \rightarrow 2^{\mathcal{Y}}$ is said to satisfy \textit{exact conditional coverage} if $$\mathbb{P}(Y \in \mathcal{C}(X) | X = x) = 1 - \alpha \mbox{ for all }x \in \mathcal{X}. $$ 

Exact Conditional coverage is known to be impossible to achieve for conformal prediction \cite{cp_imposs}; it is nonetheless desirable to achieve approximate conditional coverage. 

Firstly we consider a metric introduced specifically for classification problems in \cite{angelopoulos2020sets}, namely Size-Stratified Coverage Violation (SSCV) metric. Let $\mathcal{S} = \lbrace \mathcal{S}_i \rbrace_{i=1}^{s}$ be a disjoint stratification of the possible prediction set sizes such that $\bigcup_{j=1}^{s} \mathcal{S}_i = \lbrace 1, \dots, | \mathcal{Y} | \rbrace$ (note that each stratum may contain several possible sizes; this is useful in the case where there are many classes). Define the index set of all the datapoints with prediction set size falling into stratum $\mathcal{S}_j$ as $\mathcal{I}_j = \lbrace i : | \mathcal{C}(X_i)| \in \mathcal{S}_j \rbrace.$ Then the SSCV is defined as 
\begin{multline} \label{eq:sscv}
    \operatorname{SSCV}\left(\mathcal{C}, \mathcal{S} \right):= \\ \sup _{\mathrm{j}}\left|\frac{\left|\left\{\mathrm{i}: \mathrm{Y}_{\mathrm{i}} \in \mathcal{C}\left(\mathrm{X}_{\mathrm{i}}\right), \mathrm{i} \in \mathcal{I}_{\mathrm{j}}\right\}\right|}{\left|\mathcal{I}_{\mathrm{j}}\right|}-(1-\alpha)\right|.
\end{multline}

Here prediction set size can be thought of as a proxy for the difficulty of the sample, so intuitively this measures the worst case coverage violation conditioned on the difficulty of the example. 

In the graph setting, a desirable property for a conformal prediction method is that the coverage probability for a given node does not depend on its location within the graph. To give a trivial example of why this is important, suppose our test network consists of 100 nodes split into two communities, with 90 nodes in one community and 10 nodes in the other. To achieve $90\%$ coverage, a conformal prediction method could predict the full label set on all nodes in the first community and the empty set on all the nodes in the second community, yet this would not be helpful in practice. 

To measure the degree to which this issue is present in the constructed prediction sets, we introduce the Partition Conditional Coverage Violation (PCCV) metric. Given a partitioning of the node set into disjoint partitions $\mathcal{V} = \lbrace \mathcal{V}_i \rbrace_{i=1}^{N}$, we define the PCCV as 
\begin{multline} \label{eq:NCCV}
    \operatorname{PCCV}\left(\mathcal{C}, \mathcal{V} \right):= \\ \sup _{\mathrm{j}}\left|\frac{\left|\left\{\mathrm{i}: \mathrm{Y}_{\mathrm{i}} \in \mathcal{C}\left(\mathrm{X}_{\mathrm{i}} \right), \mathrm{i} \in \mathcal{V}_{\mathrm{j}}\right\}\right|}{\left|\mathcal{V}_{\mathrm{j}}\right|}-(1-\alpha)\right|.
\end{multline}
We estimate the SSCV using the experimental setup introduced in Section \ref{sec:e_setup}. We modify the experimental setup slightly for computing PCCV, instead taking half of the nodes (chosen at random) as calibration data and the other half as test data. We partition the data into disjoint neighbourhoods using a recursive approach; given the current set of test nodes, we choose a root node at random and take the 2-hop neighbourhood around it, then delete all the nodes in the neighbourhood from the current set and repeat until there are no neighbourhoods of size 30 nodes or more. We repeat this procedure 100 times and report the median over the different runs.

{\color{black}{The experimental results are shown in Table \ref{tab:nccv}. All NAPS variants outperform APS on all data sets and models. Within the NAPS variants, the linear weights tend to perform best on the Reddit2 data set, and the geometric weights perform very well on the Amazon dataset, while the unweighted NAPS version performs  well on all data sets.}}

\begin{figure*}[ht]
    \centering
    \includegraphics[width=\textwidth]{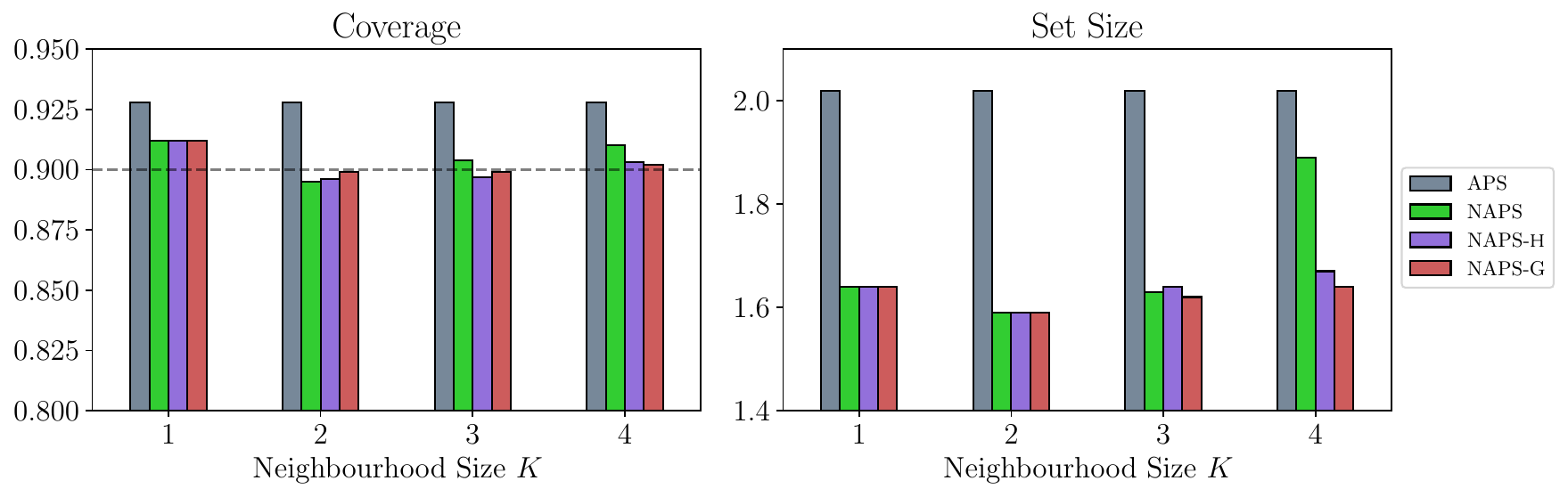}
    \caption{The coverage and size for the different conformal prediction procedures whilst varying $K$ on the Reddit2 dataset. All the experiments use the GraphSAGE-Mean GNN architecture. Each column shows the median of means over 100 repetitions of the experiment for the given method using the methodology introduced in Section \ref{sec:e_setup}. The dashed black line shows the desired coverage level $\alpha = 0.9$. }
    \label{fig:ablation}
\end{figure*}

\section{Ablation Study} \label{sec:ablation}
The neighbourhood depth parameter $K$ reflects a trade-off between sample size and similarity between datapoints. To better understand the influence of neighbourhood depth on the performance of NAPS, we repeat the Reddit2 experiment using the same setup introduced in Section \ref{sec:e_setup} for $K = 1, \dots, 4$.

The results for the GraphSAGE-Mean architecture are displayed in Figure \ref{fig:ablation}. We see that for $K=1$ (on which all three NAPS variants are equal) our method slightly overcovers, likely due to the low sample size. We see that as $K$ increases (i.e. the neighbourhoods become large) the performance of the unweighted variant of NAPS tends towards that of APS as expected. The weighted variants are better able to manage the trade-off between sample size and relevance, maintaining similar rates of coverage and prediction set size as the neighbourhood size grows.

\section{Conclusion and Future Work} \label{sec:conc}
In this work we have introduced NAPS, an approach for constructing prediction sets on graph structured data. NAPS is quick to train and deploy and comes with theoretical guarantees on the coverage. We have shown through extensive experiments that NAPS produces prediction sets that are both more efficient and have better conditional coverage properties than a naive application of conformal prediction.

In this work we have treated the neighbourhood size $K$ as a hyper-parameter. Intuitively, one would like to select this value such that the calibration nodes are "local" within the network while providing a large enough sample size to accurately estimate the quantile $1 - \hat{\alpha}$. It would be useful to further study the interplay between the optimal choice of $K$, the homophily level and the diameter of the network. NAPS could also be extended to heterophilic networks; in a heterophilic network nodes tend to be connected to dis-similar nodes. One could therefore calibrate among alternating neighbourhoods $\bigcup_{j=1}^{k} \mathcal{N}^{2j}_{n+1} \setminus \mathcal{N}^{2j-1}_{n+1}$.

{\color{black}NAPS may produce wide prediction sets when deployed on low density networks as the sample size for conformal calibration will be small, see Equation \eqref{eq:lower_upper}}. An approach for conformal prediction in hierachical models was introduced in \citet{hierachical_cp}, where the quantiles are calibrated in different groups before being pooled. An approach similar to this could be applied for nodes in small connected components, where calibration on similar neighbourhoods or components could be pooled to provide a better estimate of the conformal quantile. 

Finally, NAPS could be extended to model other types of graph structured data. Here we have only considered unweighted and undirected networks, but {\color{black}{addressing both weighted and directed networks}} %both 
follow as natural extensions to the method. It would also be fairly straightforward to extend the method to node \textit{regression} tasks simply by using a different conformal score function such as CQR \cite{cqr}. 
% \section*{Acknowledgements}
% The author would like to thank Gesine Reinert and Stefanos Bennett for their feedback on earlier versions of this work. The author is supported by the EPSRC CDT in Modern Statistics and Statistical Machine Learning (EP/S023151/1) and the Alan Turing Institute (EP/N510129/1).

{\color{black}
\section{Ethical Concerns}
While we do not believe this work is likely to have any negative impact on society, as always with statistical methods care must be taken with the interpretation of the results. Just like any method that constructs prediction intervals, those obtained by NAPS indicate a statistical prediction, and as such if used in a high-stakes field such as healthcare must be interpreted as such.
}
% For natbib users:
\bibliographystyle{unsrtnat}
\bibliography{reference}
% For bibLaTeX users:
% \printbibliography

\appendix
\section{Appendix}

{\color{black}
\subsection{Proof of Lemma \ref{lemma:1}} \label{app:proof}

We define $d := d_{TV}(\pi_1, \pi_2)$ as the total variation distance between the data distributions for the two communities.  

A naive application of APS ignores the link structure, applying weight $w_j = 1$ to all nodes. For whichever node the prediction set is being constructed, $n$ nodes will belong to the opposing community (hence occur a penalty of $d$ in the Equation \eqref{eq:gapbound}) and $n-1$ nodes will belong to the same community so we may immediately apply Equation \eqref{eq:gapbound} to obtain
\begin{equation}
    E[CG_{\text{APS}}] \leq \frac{nd}{(2n-1) + 1} = \frac{d}{2}    
\end{equation}

Define $V_i := \mathbb{I}_{Y_{i} \in \widehat{C}_n(X_{i})}$ as the indicator function of the event that the $i^{th}$ node is covered by the constructed prediction set. We can write the expected coverage for NAPS as 
$$ \mathbb{E}[C_{\text{NAPS}}] = \mathbb{E}\left [ \frac{1}{2n}\sum_{i=1}^{2n} V_i \right ] = \frac{1}{2n}\sum_{k=1}^{2n} \mathbb{E}[V_k].$$

We can write each term in this sum as
% \begin{multline} \label{eq:condexpansion}
%     \mathbb{E}[V_k] = \sum_{i=1}^{n-1} \sum_{j=1}^n  \mathbb{E}[V_k \mid N_{in} = i, N_{out} = j]  \\ Bin(i; n-1, p_{in}) Bin(j;n, p_{out})
% \end{multline}

\begin{equation} \label{eq:condexpansion}
    \begin{split}
        \mathbb{E}[V_k] &= \sum_{i=1}^{n-1} \sum_{j=1}^n \mathbb{E}[V_k \mid N_{\text{in}} = i, N_{\text{out}} = j] \\
        &\quad \times Bin(i; n-1, p_{\text{in}}) \, Bin(j; n, p_{\text{out}})
    \end{split}
\end{equation}

where $N_{in}, N_{out}$ are random variables indicating the number of neighbours in the same community and opposing community to node $k$ respectively, and $Bin(l; n, p)$ is the probability mass function of a Binomial random variable with parameters $n$ and $p$ evaluated at $l$. 

The term $\mathbb{E}[V_k \mid N_{in} = i, N_{out} = j]$ is just the coverage of a prediction set constructed for node $k$ in a fixed realisation of the network, so we may apply Equation \eqref{eq:gapbound} to obtain
\begin{equation}
    \mathbb{E}[V_k \mid N_{in} = i, N_{out} = j] \geq (1 - \alpha) - \frac{jd}{i + j + 1}. 
\end{equation}

Lower bounding term-wise in Equation \eqref{eq:condexpansion} we have that 

\begin{equation}
    \mathbb{E}[V_k] \geq (1 - \alpha) - d \mathbb{E} \left [\frac{N_{out}}{N_{out} + N_{in} + 1} \right ]
\end{equation}

Comparing the two bounds gives the result.
}
\subsection{Dataset Details} \label{app:dataset}
For the experiments above we used the Flickr and Reddit2 datasets from \cite{graphsaint-iclr20}, {\color{black}and the Amazon Computers dataset introduced in \cite{pitfalls}}. 

{\color{black}The Flickr dataset is constructed using images uploaded to the Flickr site, where the node features consist of the meta-data for each image and the label is the image tag. The Reddit2 dataset is constructed from posts on the social media site Reddit, with posts representing nodes. The node features are bag-of-word vectors from the post, and the label is the community (or sub-reddit) that the post belongs to. The Amazon Computers dataset consists of segments of an Amazon co-purchase graph, where nodes represent goods and links are added between nodes if they are frequently bought together.

Our train/validation/test splits for Flickr and Reddit2 were done using the splits given in the original papers (which are conveniently implemented in Pytorch Geometric \cite{pyg}). For Amazon Computers we constructed our own split, using 752 nodes for training, 1000 for validation and the remaining 12000 for testing. As mentioned in the main text we tested our graph only on large connected components, which we chose as nodes with at least 50 2-hop neighbours in Flickr and Amazon Computers, and nodes with at least 1000 2-hop neighbours in Reddit2. We call this set of nodes $\mathcal{N}^{cal}$, and report the sizes of these sets as well as some summary statistics about each dataset in Table \ref{tab:datasets}.
}

% {\color{black}
% \subsection{Experimental Results for the Amazon Computers Dataset}\label{app:amazon}
% Here we present the results for the Amazon Computers dataset as described in Section \ref{app:dataset}. The results are displayed in Table \ref{tab:Amazon}; NAPS produces tighter prediction sets than APS in all cases and in most provides coverage closer to the $90\%$ level. 

% }
\subsection{Model Training Details} \label{app:training}
We used the implementations of GraphSAGE and ShaDow provided by Pytorch Geometric \cite{pyg}. All models on all datasets used the same hyper-parameters. Each GNN used 2 layers with hidden dimension $H=64$. We used the Adam optimiser \cite{adam} with default hyper-parameters, learning rate $\eta = 0.1$, and used dropout probability $\delta = 0.5$. For the GraphSAGE neighbour sampling training we used 25 1-hop neighbours and 10 2-hop neighbours. We used early stopping based on the accuracy on the validation set. We made no effort to optimise any of these parameters as we are not trying to optimise for accuracy, {\color{black}{we merely assess whether}} our method performs well with a variety of architectures.  \

Each experiment here took less than two hours in total on a single machine with an NVIDIA GeForce RTX 2060 SUPER GPU and an AMD Ryzen 7 3700X 8-Core Processor. One run of the conformal prediction procedure has trivial overhead when compared with model fitting (and actually NAPS is faster than APS as we use less data points to calibrate the procedure).

\end{document}